\pdfoutput=1
\documentclass{article}

\usepackage[preprint]{neurips_2020}
\usepackage{graphicx}
\usepackage[utf8]{inputenc} 
\usepackage[T1]{fontenc}    
\usepackage{hyperref}       
\usepackage{url}            
\usepackage{booktabs}       
\usepackage{amsfonts}       
\usepackage{nicefrac}       
\usepackage{microtype}      
\usepackage{stackengine}
\usepackage{multirow}

\def\delequal{\mathrel{\ensurestackMath{\stackon[1pt]{=}{\scriptstyle\Delta}}}}

\newcommand{\cross}[1][1pt]{\ooalign{%
  \rule[1ex]{1ex}{#1}\cr
  \hss\rule{#1}{.7em}\hss\cr}}

\title{Generating Out of Distribution Adversarial Attack using Latent Space Poisoning}

\author{%
  Ujjwal Upadhyay\\
  Department of Computer Science\\
  Bharati Vidyapeeth's College of Engineering\\
  New Delhi, Delhi 110063 \\
  \texttt{ujjwalupadhyay8@gmail.com} \\
   \And
   Prerana Mukherjee \\
   Department of Electrical Engineering\\
   IIT Delhi \\
   New Delhi, Delhi 110016 \\
   \texttt{mukherjee.prerana@gmail.com} \\
}

\begin{document}

\maketitle

\begin{abstract}
Traditional adversarial attacks rely upon the perturbations generated by gradients from the network which are generally safeguarded by gradient guided search to provide an adversarial counterpart to the network. In this paper, we propose a novel mechanism of generating adversarial examples where the actual image is not corrupted rather its latent space representation is utilized to tamper the inherent structure of the image while maintaining the perceptual quality intact and to act as legitimate data samples. As opposed to gradient-based attacks, the latent space poisoning exploits the inclination of classifiers to model the independent and identical distribution of the training dataset and tricks it by producing out of distribution samples. We train a disentangled variational autoencoder ($\beta$-VAE) to model the data in latent space and then we add noise perturbations using a class-conditioned distribution function to the latent space under the constraint that it is misclassified to the target label. Our empirical results on MNIST, SVHN, and CelebA dataset validate that the generated adversarial examples can easily fool robust $l_0$, $l_2$, $l_{\infty}$ norm classifiers designed using provably robust defense mechanisms.
\end{abstract}

\section{Introduction}

Artificial Neural Networks (ANN's) are vulnerable to adversarial examples, even an insignificant change to human eyes might still fool even the state of the art classifier \citep{Szegedy2014IntriguingPO}. This motivates the exploration of adversarial learning domain of artificial intelligence to make the neural networks more robust and secure. Deep learning models are exposed to all kind of threats like \citep{DBLP:journals/corr/abs-1802-06430} \citep{DBLP:journals/corr/abs-1907-00374} \citep{8844593} demonstrates the need for further strengthening the models and removing bias towards the presented training dataset distribution. To train the model against such threats, many methods have been developed. Most of these basically try to exploit adversarial training of the model \citep{Goodfellow2015ExplainingAH} \citep{DBLP:conf/iclr/MadryMSTV18} \citep{Sinha2017CertifiableDR}. Some other techniques aim to remove the adversarial perturbation \citep{DBLP:journals/corr/abs-1712-02976} \citep{song2018pixeldefend}.

In this paper, we introduce a novel method to generate adversarial examples that are smooth by nature and provide realistic perception. In practise adversarial examples are generally noisy in nature and are easily traceable by humans. They can also be mitigated by above mentioned defense techniques. So in this work, we make use of disentanglement properties of $\beta$-VAE networks \citep{Higgins2017betaVAELB} to get the meaningful latent space vector which can be tuned with visible results in image space. We utilize a specialized variant of $\beta$-VAE network namely \textit{JointVAE} \citep{dupont2018learning} which provides more control over the latent space. We conditioned our proposed model such that, it promotes the class with second highest probability of occurrence and makes it visually appear in the image or just change the classification prediction. We perform this by using statistical theory concepts and adding the distribution of that class, extracted from training set, to the actual latent space vector of the original input image. This not only gives smoothing effect to the image but also leverages additive property of latent space and linear interpolation smoothness produced by such operations. Then we employ our pre-trained JointVAE model to decode the new noisy latent space and obtain adversarial examples conditioned under some constraints, mentioned in section~\ref{section:methods}. The other contribution of our attack is that it is completely black-box attack i.e it doesn't require any access to the test classifier.

\section{Background}
In this section, we discuss the background on adversarial examples, attacking methods, defense methods and also autoencoder networks. Although the topic of adversarial attack and defense ranges from image classification, to audio classification to many ubiquitous domains, however here we focus on image classification.

\textbf{Adversarial Attacks} Let $x \in \mathbb{R}^n$ and $f$ be a classifier. If $f$ is known to attacker then it is called white-box attack, if it is not known then it is called black-box attack. In white-box attacks, matrix norms such as $l_0$ \citep{7467366}, $l_2$ \citep{7780651}, or $l_{\infty}$ \citep{Kurakin2017AdversarialML}  are used in combination with gradient based methods like Fast Gradient Sign Method (FGSM \citep{Szegedy2014IntriguingPO}), which assumes image or input itself to be updated with incoming gradient and thus changes the pixel values accordingly. If this procedure is repeated several times, then it is termed as Projected Gradient Descent (PGD \citep{journals/corr/KurakinGB16}) attack. 

\textbf{Adversarial defenses} Prevalent adversarial defenses focuses on making classifier robust in response to perturbations in the images. These emphasize on adversarial training with adversarial examples added to the training set. \citep{DBLP:journals/corr/abs-1802-00420} indicates that all the existing defenses depend on certified defense or adversarial training \citep{DBLP:conf/iclr/MadryMSTV18}. Although some recent certified defense techniques have emerged which finetune the robust model in their latent space \citep{DBLP:journals/corr/abs-1905-05186} which is different from traditional image space based adversarial training. In this paper, we propose an attack to fool the adversarial defense trained using adversarial training. 

\textbf{Autoencoder Networks} An autoencoder has 2 blocks, an encoder network, $E$ and other decoder network, $D$. The encoder network translates the actual image in pixel space to the latent space, whereas the decoder network translates the latent space representation of the image back to pixel space. There is some loss during this translation but the integrity of the image remains intact in terms of its structure and some prevalent details. So, the goal of autoencoder training is to learn the encoder and decoder coupling in such a way that the reconstruction loss is minimal. In order to get more variability in the latent space, we use JointVAE \citep{dupont2018learning} network, which gives us more control over the induced changes in the pixel space. It is optimized using following optimization problem,
\begin{equation}
\label{eq:optimization}
    \mathcal{L}(\theta, \phi) = \mathbb{E}_{q(z,c|x)}[log p_{\theta(x|z,c)}] - \beta D_{KL}(q_{\phi}(z,c|x) || p(z,c))
\end{equation}
where, the joint posterior $q_{\phi}(z,c|x)$ is mapped by encoder network, Gaussian prior $p(z,c)$ and likelihood $p_{\theta}(x|z,c)$ is mapped by decoder network. Here latent distribution is jointly continuous and discrete with the assumption that both continuous ($z$) and discrete ($c$) variables are conditionally independent.

\section{Methods}
\label{section:methods}

\subsection{Out of Distribution Examples}
\textbf{Traditional gradient based attacks} or perturbation-based attacks work on the principle of back-propagation where we try and calculate the gradient for each input pixel. This gradient is used to define a perturbation vector under a constraint that the adversarial image is similar to the original image.\\\\
\textbf{Definition 1} (Gradient Based Attacks). Given a subset of test images $ \tau \subset o $, where $o$ is the whole dataset distribution, small constant $ \epsilon > 0 $, and matrix norm $||.||$, a perturbation vector is defined. The adversarial image, $x'$ constructed by this perturbation vector is defined to be any image in $\alpha$ as,
\begin{equation}
\label{eq:perturbation}
  \alpha \delequal \{x \in o \ | \ \exists x' \in \tau, \ ||x-x'|| \leq \epsilon \wedge f(x') = \theta(x') = \theta(x) \neq f(x)\}  
\end{equation}
where $\theta$ is a human evaluator, $f$ is a classifier and $\mathcal{I}$ is any image. 
\begin{equation}
\label{eq:generalclassifier}
 f: \mathcal{I} \rightarrow \{1,2,...,K\}  
\end{equation}

In our work, \textbf{Out of Distribution Adversarial Attack using Latent Space Poisoning} we present out of distribution images caused by latent space poisoning, which follows the below definition.\\\\
\textbf{Definition 2} (Out of Distribution Adversarial Attack). An out of distribution example is any image that is element of $\beta \delequal \{x \in o \ | \ o(x) \neq f(x)  \}$

In prior works, $\theta$ is chosen to be a human evaluator, $\tau$ is test data, and $||.||$ is a matrix norm which is either $l_{\infty}$ or $l_{2}$ norm. So, $o$ should represent the images which seems realistic to the human evaluator. As the assumption of traditional attacks, we also adhere to the congruent lines that new image $x'$ should be similar to original image $x$ but with distinction of working in latent space rather than image space itself. From this, it should be clear that $\alpha \subset \beta$. This means our proposed attack is a generalization of traditional attacks in a more flexible manner where latent vector can be used to regenerate a smooth image in image space without much visible noise which was earlier controlled with $\epsilon$. The work over latent space provides smoothing effect to the human evaluator $\theta$.

\subsection{Practical Attack using Latent Space Poisoning}

The key to produce adversarial attacks that are out of distribution of training set is to model the set of legitimate images in the latent space. We do so by training a JointVAE $A$, a derivative of variational autoencoders \citep{kingma2014autoencoding}. This is done to map latent space in $z \in \mathbb{R}^{m}$, a set of continuous latent space variables, and $c \in \mathbb{R}^{n}$, a set of discrete latent variables. This gives meaning to our latent space in a sense that each latent factor controls a unique aspect about the original image. In practice, such a latent space is able to model the training data using a fairly accurate decoder which models the likelihood i.e. $P_{\theta}(x|z,c)$. Here $z$ and $c$ are conditionally independent.

Then a classifier model $C_1$ is chosen, different from the actual classifier $C_2$ put under test, to classify the generated image, $C_1(x')$. This classification result produces soft labels giving us information about similarity of images with other classes in the dataset. For example, if a classifier produces soft labels to be $\{ \mathcal{A}, \mathcal{B}, \mathcal{C} \} \rightarrow \{ 0.7, 0.2, 0.1 \}$. Then it means that classifier predicts or models the current input as $\mathcal{A}$ to be highly probable with probability of $0.7$, $0.2$ for $\mathcal{B}$ and $0.1$ for $\mathcal{C}$. This gives us an intuition about the actual modelling of the classifier. The assumption about this pre-trained classifier is that it should be able to classify with high accuracy and low loss. This makes it certain that modeling of the test classifier will be more or less similar to our trained classifier when used in combination with label smoothing \citep{DBLP:journals/corr/abs-1906-02629}. This information about soft logits gives us meaningful insights about the data distribution and we can exploit it by perturbing noise in the latent space in order to make the soft labels deviate from actual target label in a subtle way. The minute change to the latent space in our proposed work is to make second most probable label to have equal probability as the class with highest probability.\\

\textbf{Understanding Latent Space Poisoning}

Let $A_{\theta}(x)$, $C_{\phi}(x)$ be the variational autoencoder and pre-trained inference classifier respectively. Let $f(x)$ be the classifier we want to attack. Here, we put our focus on out of distribution adversarial attack which generate a target image $x'$ where $o(x') \neq f(x')$, where $o(.)$ is a human evaluator. In order to produce the adversarial examples of this sort, we propose finding an appropriate latent space vector, $z'$ (continuous factors) and $c'$ (discrete factors). This is done by adding some noise to the predetermined latent space vectors of the input in their continuous and discrete factors. This is done by firstly finding the distribution of latent space vector for each class in the training set.
\begin{equation}
\label{equation:distribution-latent-space}
    \mathcal{D}(y_i) \rightarrow \{(\mu_{z}^{y_i}, \sigma_{z}^{y_i}), (\mu_{c}^{y_i}, \sigma_{c}^{y_i})\} \; \forall \, i \in \mathcal{O}
\end{equation}

where $\mathcal{O}$ denotes classes. This distribution function $\mathcal{D}$ helps in generating a latent space vector which when added to the original latent space along with some noise $\eta$, produce the desired result regarding the output image. We have utilized additive noise over multiplicative noise because \citep{kingma2014autoencoding, Ozair2014DeepDG}
have shown in their work that the linear interpolation of images in latent space is smoother and realistic and is more likely to be similar to the training distribution.
\begin{equation}
\label{equation:z-latent-creation}
    z' = \mu_{z} + \eta_{z} \cdot \sigma_{z}
\end{equation}
\begin{equation}
\label{equation:c-latent-creation}
    c' = \mu_{c} + \eta_{c} \cdot \sigma_{c}
\end{equation}

The aforementioned noise vector is then produced by a neural network, given the actual latent space produced by variational autoencoder. The produced latent space, $z'$ and $c'$ are then added to the original latent space vector, 
\begin{equation}
\label{equation:nosied-z-latent-creation}
    \mathcal{Z} = \lambda_{org} \cdot z + \lambda_{noised} \cdot z'
\end{equation}
\begin{equation}
\label{equation:nosied-c-latent-creation}
    \mathcal{C} = \lambda_{org} \cdot c + \lambda_{noised} \cdot c'
\end{equation}

where $\lambda_{org}$ and $\lambda_{noised}$ are hyperparameters. Then $\mathcal{Z}$ and $\mathcal{C}$ are decoded and an adversarial output is produced. These latent space vectors, $\mathcal{Z}$ and $\mathcal{C}$, help in producing out of distribution sample for a particular class because the point estimates for each class is independent of other classes. Hence combination of these distributions will output a distribution with different point estimates.
\begin{equation}
    \mathcal{X} = \mathcal{N}(\mu_{x}, \sigma_{x}^2) \;\;\; , \;\; \mathcal{Y} = \mathcal{N}(\mu_{y}, \sigma_{y}^2)
\end{equation}
\begin{equation}
    \mathcal{T} = X + Y
\end{equation}
\begin{equation}
    \mathcal{T} \sim \mathcal{N}(\mu_{x} + \mu_{y}, \sigma_{x}^2 + \sigma_{y}^2)
\end{equation}
The original dataset distribution is the subset of resulting distribution from where adversarial examples are sampled. $\mathcal{Y}$ is the original dataset distribution.
\begin{equation}
    \mathcal{Y} \subset \mathcal{T}
\end{equation}
Now this process of generating adversarial samples is optimized by minimizing the loss function $\mathcal{L}$. This loss function $\mathcal{L}$ is designed in such a way that the generated image keeps resemblance with the original image and minimal pixels are manipulated significantly. This helps in producing high fidelity adversarial examples. The loss function is a combination of different constraints that affects the target image.
\begin{equation}
    \mathcal{L} \delequal \lambda_{0} \mathcal{L}_0 + \lambda_{1} \mathcal{L}_1 + \lambda_{2} \mathcal{L}_2 
\end{equation}
where $\lambda_0$, $\lambda_1$ and $\lambda_2$ are the hyperparameters. The first cost $\mathcal{L}_0$ has $\widehat{x}$ as the predicted soft labels and $\widehat{y}$ is the target soft labels. We manipulate predicted $\widehat{x}$ in such a way that the probability for second most probable label in the input sample becomes more than or equal to the most probable label in actual prediction from classifier. So, $P_{\phi}(\widehat{x}^{2}) \geq P_{\phi}(\widehat{x}^{1})$
\begin{equation}
\label{equation:BCE}
    \mathcal{L}_0 = \sum -w_{n}[y_n \cdot \log(x_n) + (1-y_n) \cdot \log(1-x_n)]
\end{equation}
The first cost function encourages that the label with second highest probability is somehow induced as some pixel changes in the original input such that the classifier changes its predictions and gets confused between the original class and adversarial class.\footnote{Adversarial class here means the class that we intend the classifier to predict for the given input with minimal changes.}. The equation~\ref{equation:BCE} is a Binary Cross Entropy (BCE) loss which is usually used in multi-label classification. It helps in inferring when there are 2 classes present in the input image. In similar way, it helps to get the essence of 2 classes, one originally present and other induced class (as a result of adversarial example), in the same input image.

The second cost function can be any of these two functions given below, 
\begin{equation}
    \mathcal{L}_1 = ||x-x'||_{2} \;\;\;or \;\;\; ||x-x'||_{\infty}
\end{equation}
It encourages that the generated image is as close to the original image as much as possible.\\\\
The third cost function is the structural dissimilarity index, 
\begin{equation}
    \mathcal{L}_2 = 1 - SSIM(x,x')
\end{equation}
\begin{equation}
    SSIM(x,x') = \frac{(2u_{x}u_{x'}+c_1)(2\sigma_{xx'}+c_2)}{(u_x^2+u_{x'}^2+c1)(\sigma_x^2 + \sigma_{x'}^2 + c_2)}
\end{equation}
\begin{equation}
    c_1 = (k_1L)^2, \;\; c_2 = (k_2L)^2, \;\; L = 2^{\mbox{bits per pixel}} -1
\end{equation}
where \(u_x\), \(u_{x'}\) are the average of all the pixel values of the image \(x\) and \(x'\) respectively. \(\sigma_x^2\), \(\sigma_{x'}^2\) are the variance of the pixel value of the image \(x\) and \(x'\) respectively. \(\sigma_{xx'}\) is the covariance between \(x\) and \(x'\). \(c_1\) and \(c_2\) are the two variables to stabilize the division with weak denominator. \(L\) is dynamic range of pixel values. 
It encourages that the spatial correlation among pixels is maintained. Hence imposing minimal change between adjacent pixels and thereby limiting the case of abrupt changes in nearby pixels.

\section{Experiments}

\textbf{Datasets} The datasets used in the experiments are MNIST \citep{10.1162/neco.1989.1.4.541}, CelebA \citep{7410782}, and SVHN \citep{Netzer2011ReadingDI}. Out of these, MNIST and SVHN are digits dataset, while CelebA is face dataset where we intend to classify images based on gender. We attack such classification tasks because of less complicated attributes to identify (such as digit labeling, gender classification) in verifying the results of classifier and to enable labeling of the generated image by our attack.

\textbf{Model setting} We train JointVAE \citep{dupont2018learning} on all the datasets. We only trained the JointVAE on training and validation data partition of each dataset. We utilized test dataset for testing our classifier that we trained to get inference regarding the similarity between different classes. We have utilized ResNet \citep{7780459} block in classifier model for CelebA and SVHN dataset while we used Madry architecture \citep{DBLP:conf/iclr/MadryMSTV18} for MNIST classifier. 


\subsection{Untargeted attack against certified defenses}

We here validate that our adversarial attack can bypass certified defense \citep{DBLP:conf/iclr/MadryMSTV18}. These defenses show that with a pre-defined perturbation less than $\epsilon$, a training example won't be classified incorrectly.

\textbf{Setup} For each test sample from test partition, we run our attack and generate an adversarial example for that. Then these generated samples are fed to the target classifier and their predictions are recorded.

\textbf{Results} Table~\ref{results} shows the accuracy of the test classifier after applying our attack. And since our attack is not image space perturbation attack rather it is latent space perturbation, it is able to fool the classifier for around $46\%-54\%$ of the samples for MNIST. Similar observations can be inferred for other datasets as well. Table~\ref{ablation:size} shows the effect of network size, \textit{total parameters = parameters(noise learner) + parameters(coefficient learner)}, where the change in network size or trainable parameters represents changing the noise $\eta$ learning model and coefficients learning model (learning how much noise to transfer to the tuning parameter $\beta$, which gives further control over noise addition in latent space). It can be inferred from Table~\ref{ablation:size} that number of trainable parameters doesn't affect the success rate of the attack after a threshold which in MNIST case is around 12K parameters. The effect of $\alpha$ and $\beta$ hyperparameters can also be seen from the result where it can be clearly seen that decreasing $\alpha$ value increases the success rate or decrease the accuracy of the test classifier. $\cross$ denote the best public results.

\begin{table}[h!]
  \caption{Results on CelebA, MNIST, SVHN}
  \label{results}
  \centering
  \begin{tabular}{llll}
    \toprule
    Datasets     & Architecture     & Success rate of our attack & Success rate of PGD attack$^{\cross}$ \\
    \midrule
    MNIST & Madry Network \citep{DBLP:conf/iclr/MadryMSTV18}  & $\sim 50\%\pm 4\%$  & $10.4\%$ ($\epsilon=0.3$)\\
    CelebA     & WideResNet \citep{BMVC2016_87} & $\sim 46\% \pm 5\%$ &  $59.9\%$ ($\epsilon=0.03$)    \\
    SVHN     & WideResNet  \citep{BMVC2016_87}     & $\sim 80\% \pm 6\%$  & $20.5\%$  ($\epsilon=0.03$) \\
    \bottomrule
  \end{tabular}
\end{table}

\begin{table}[h!]
  \caption{Results on MNIST with hyperparameter tuning}
  \label{ablation:size}
  \centering
  \begin{tabular}{llll}
    \toprule
    $\alpha$ & $\beta$ & Trainable Parameters & Accuracy \\
    \midrule
    0.6 & 0.8  &    12K           & $\sim 70\%\pm 4\%$      \\
    0.5 & 0.8  &    12K           &$\sim 69\% \pm 3\%$      \\
    0.3 & 0.8  &    12K           &$\sim 67\% \pm 5\%$      \\
    0.1 & 0.8  &    12K           &$\sim 60\% \pm 4\%$      \\
    0 & 1      &    12K           &$\sim \textbf{50}\% \pm \textbf{4}\%$      \\
    0.6 & 0.8  &    29K           &$\sim 70\% \pm 3\%$      \\
    0.5 & 0.8  &    29K           &$\sim 68\% \pm 3\%$      \\
    0.3 & 0.8  &    29K           &$\sim 67\% \pm 3\%$      \\
    0.1 & 0.8  &    29K           &$\sim 58\% \pm 4\%$      \\
    0 & 1      &    29K           &$\sim 60\% \pm 3\%$      \\
    \bottomrule
  \end{tabular}
\end{table}

Table~\ref{test:classifier} shows the actual accuracy of classifier being used for the experiments. Table~\ref{test:classifier} shows the robustness and standard accuracy of the classifier where robustness accuracy tells the accuracy of classifier trained with adversarial examples. Table~\ref{test:classifier} also shows inference classifier used by our training architecture to know the resemblance of input image with other classes as perceived by our classifier.

\begin{table}[h!]
  \caption{Accuracy of different classifiers, inference and test classifier}
  \label{test:classifier}
  \centering
  \begin{tabular}{llll@{\hspace{3em}}ll}
    \toprule
        \multirow{1}{*}{} &
      \multicolumn{3}{c}{Test Classifier (Attacked)} &
      \multicolumn{2}{c}{Inference Classifier} \\
    \cmidrule(r){2-4}
    \cmidrule(r){5-6}
    Datasets & Architecture     & Standard Accuracy  & Robust Accuracy & Architecture & Accuracy \\
    \midrule
    MNIST & Madry Network \citep{DBLP:conf/iclr/MadryMSTV18}  & $98.7\%$ & $98.4\%$    & LeNet \citep{726791}  & $97.5\%$       \\ 
    CelebA     & WideResNet \citep{BMVC2016_87} & $96\%$ & $96\%$   & ResNet152 \citep{7780459} & $97.5\%$      \\ 
    SVHN     & WideResNet \citep{BMVC2016_87}& $86.45\%$ & $85.24\%$ & ResNet152 \citep{7780459} & $93.8\%$\\
    \bottomrule
  \end{tabular}
\end{table}

\subsubsection{Realistic perception of adversarial samples}

\begin{figure}[ht!]
    \centering
    \includegraphics[scale=0.44]{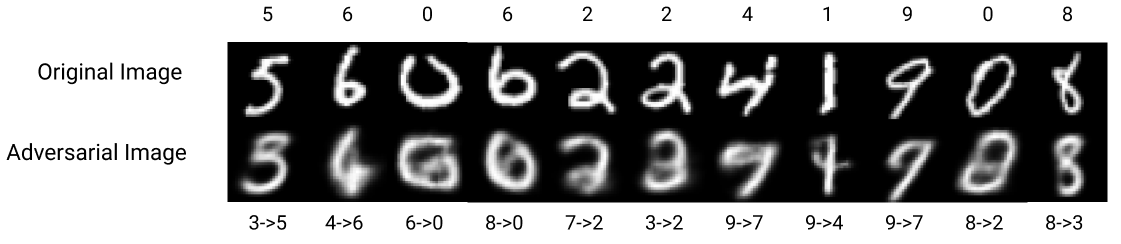}
    \caption{Figure illustrates some adversarial example generated from our attack on MNIST given the original image. The label above figure gives the actual prediction of classifier given input image, and label below gives the prediction of humans followed by '->' with classifier prediction. }
    \label{fig:mnist-results}
\end{figure}

\begin{figure}[ht!]
    \centering
    \includegraphics[scale=0.42]{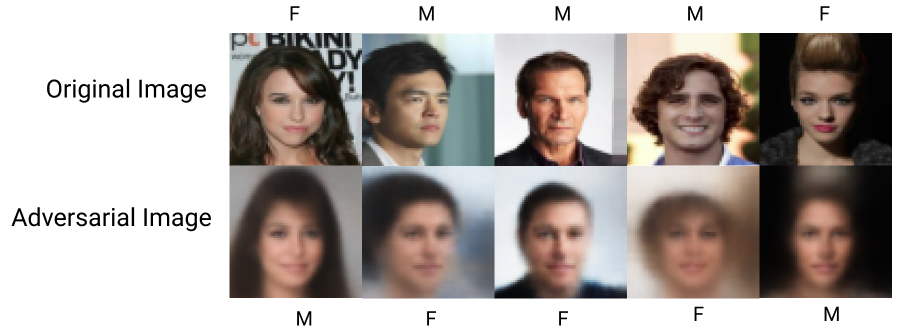}
    \caption{Figure illustrates some adversarial example for CelebA dataset. The label above image gives actual target label and label below the image is the predicted by classifier.}
    \label{fig:celeba-results}
\end{figure}

\begin{figure}[ht!]
    \centering
    \includegraphics[scale=0.42]{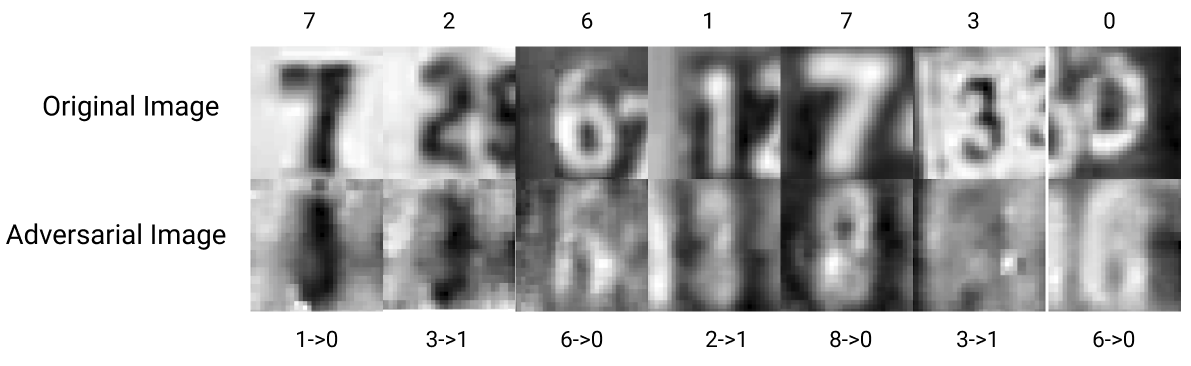}
    \caption{Figure illustrates some adversarial example for SVHN dataset. The label above image gives actual target label, and label below gives the prediction of humans followed by '->' with classifier prediction.}
    \label{fig:svhn-results}
\end{figure}

Most of our generated samples has realistic perception of some digit and this is due to the pre-trained JointVAE autoencoder used as the decoder in the model always ensures that latent space is translated to slightly manipulated image space than original image. This makes the images generated by our attack realistic and can fool human evaluators in the loop. Figures~\ref{fig:mnist-results}, \ref{fig:celeba-results}, and \ref{fig:svhn-results} are representative of qualitative results.

\subsection{Comparision with PGD}



We compare traditional PGD based attacks with our proposed attack. The PGD is a 20-step PGD with values of $\epsilon$ given in the Table~\ref{results}. It should be noted that these perturbation results are not comparable to ours because of the noisy nature of the attack i.e. if we increase $\epsilon$ to a high noise value it becomes more and more visible to humans that the image is tampered. But in our attack, if we increase the noise factor $\eta$ then we are only changing the image to a different image which also seems realistic because of the nature of translation in the latent space.

We also considered newly proposed GAN based attacks as well for some reference. So, for comparison to GAN based attacks, \citep{NIPS2018_8052} is chosen to be targeted attacks, as their results were publicly available for the datasets, we evaluated our results on. As per \citep{NIPS2018_8052}, it achieves around $84\%$ accuracy over all datasets on which we evaluated our results on. It is also important to note that these attacks generate images susceptible for attack from scratch and hence share no resemblance with the actual test images.

\section{Analysis}
\label{sec:analysis}

\begin{figure}[ht]
    \centering
    \includegraphics[scale=0.39]{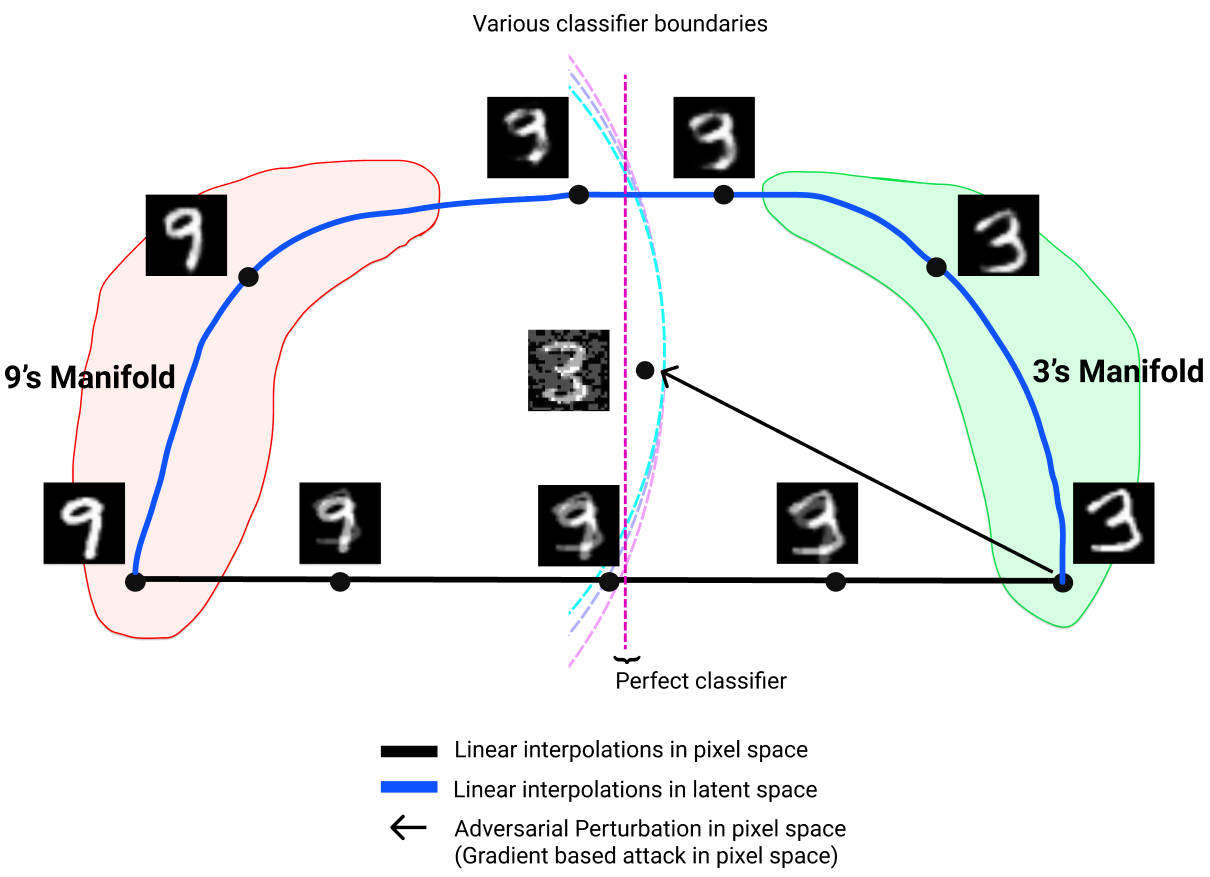}
    \caption{Illustration of working of classifier in pixel space and latent space. Smoothing effect of the latent space during interpolation is observed in the figure with contrasting difference in the adversarial images with respect to interpolation in pixel space or image space. }
    \label{fig:classifier-latent}
\end{figure}

In this section, we analyze why our threat model causes the classifier model to misclassify the adversarial samples. As shown in Figure~\ref{fig:classifier-latent}, the gradient based attacks learns the noise in the image space that will translate the adversarial image to cross the decision threshold of the classifier. But in our attack, we try and add a learnt vector to the latent space representation of the attacked image that makes its similarity and look more similar to what we intend it to classify. Since this addition of noise is performed in latent space, the image gets a smoothing effect. Our attack makes use of manifold unfolding \citep{Ozair2014DeepDG}. The autoencoder unfolds the manifolds and makes the resultant linear combination of latent space vector look realistic and is more likely to fall in original dataset distribution. The figure~\ref{fig:classifier-latent}, gives the idea regarding the realistic translation.

If $x_1$ and $x_2$ are the data points in image space of data distribution $\mathcal{Q}$ and $f$ is encoding function and $g$ is decoding function, the points on the line $h_{\alpha} = \alpha f(x_1) + (1-\alpha) f(x_2)$ yields a latent space vector which when decoded $g(h_{\alpha})$ looks like a sample from original distribution $\mathcal{Q}$. We perform our attack in a similar fashion, we try and create a vector which makes the original latent vector of original image come close to  decision threshold between the source image and target distribution.\footnote{We refer to the target space as distribution as there are many samples in a particular point which is across the decision threshold and this is controlled by $\alpha$ and $\beta$ hyperparameters. These hyperparameters clarifies regarding how aggressively the original latent vector is translated to the target distribution.} In our attack as mentioned in equation \ref{equation:distribution-latent-space}, instead of sampling another image $x_2$, we create a vector using equation \ref{equation:z-latent-creation} and \ref{equation:c-latent-creation} and subsequently translate it using,
\begin{equation}
    h_{\alpha\beta} = \alpha f(x_1) + \beta f(x_2)
\end{equation}
This gives us a controlling factor for adjusting the actual adversarial example in image space.

\section{Related Works}

Some recent attacks have explored the ideas beyond the traditional attacks which introduces perturbations to the image space or pixel space. \citep{zhengli2018iclr} proposes to map the input from image space to latent space using GANs \citep{10.5555/2969033.2969125} and then searching for adversarial examples around it.

Another line of works include attacks which are inclined towards generating adversarial examples from scratch or noise. They also utilize GAN for learning the latent space but without any reference point and therefore have exhaustive search space. AdvGAN \citep{DBLP:journals/corr/abs-1801-02610}, AdvGAN++ \citep{9022049}, AT-GAN \citep{DBLP:journals/corr/abs-1904-07793}, Defense-GAN \citep{defensegan} and \citep{NIPS2018_8052} are few works where latent space was learnt using GANs to carry out learning the distribution of the training set and generate adversarial examples accordingly. It is imperative to note here that these attacks are different from our work as we have utilized autoencoders in our work to ensure that the adversarial examples remain in a modified distribution where the input images' latent space are combined with target class (the one with second highest probability in prediction). This makes our adversarial sampling distribution to be hybrid of the 2 classes.

There are few adversarial attacks which makes use of autoencoders to perform the search in latent space with a reference point. \citep{DBLP:journals/corr/abs-1711-02879} proposes to fool the classifier by again training the pre-trained autoencoder to learn the multiplicative and additive noise in the latent space which could flip the actual prediction of the classifier given the same input image. It is also different from our work because they learn noise while we targeted the original image with the latent space of the class with learnt coefficients of variance and noise as shown in equations~\ref{equation:z-latent-creation}, \ref{equation:c-latent-creation}, \ref{equation:nosied-z-latent-creation}, and \ref{equation:nosied-c-latent-creation}. Secondly, their work is based on a white box attack paradigm while our work is based on black box attack. Thus, this gives our approach transferability over cases where test classifier is unknown.  

\section{Conclusion}

In this paper, we introduced a novel threat model, and proposed adversarial attack where actual manipulation takes place in latent space or hidden space. As demonstrated in experiments, this attack beats the current defenses against gradient based attacks (perturbation attacks). Our proposed attack lies in a gray area between existing perturbation based attacks and GAN based attacks. We take advantage of working in latent space in order to generate adversarial examples as done in GAN based attack but we perform this attack on original source image as done in perturbation based attacks. The widely used technique in optimizing machine learning models relies on empirical risk minimization \citep{10.5555/211359} however there is a trade-off between standard accuracy and adversarial robustness accuracy \citep{10.5555/3327345.3327409} and hence, gives the threat models some leverage against the discriminator models. Therefore our threat model suggests to create a new kind of training mechanism where importance of representation learning in latent space is given more importance when training the networks. This will not only increase the interpretability of the network but also provide inferences regarding where manual-tuning is required in the network.

\section*{Broader Impact}

In this section, we will describe the broader impact of our research. We will mention both possible positive (pros) and negative (cons) impacts of our research.

\textbf{Positive impact}  AI (Artificial Intelligence) has come a long way in achieving human level intelligence. It is not only able to perform tasks with superseding accuracy that are trivial for humans, but AI is now advancing to more complex areas like mathematical equation solving \citep{Saxton2019AnalysingMR}, playing games \citep{44806} and beating humans in it. So, now there is a greater need to find the vulnerabilities in these systems as they are being used in real life and sometimes failure caused by these systems may be fatal. Therefore, our work is precisely aimed at testing the robustness of the AI, and mitigate the vulnerabilities in current systems. The vulnerability discussed in the paper gives the researchers a scope to explore the latent space in depth with representational learning for various tasks rather than just inspecting the image space or raw data for performing different tasks. Our work also presents that latent space can help in generating examples in image space that correlate with multiple classes. Hence, such a technique can be used to develop sketches of criminals by crime and investigation agencies, where we can ask the identifier to describe the features of the culprit and exploit part based matching with pre-existing template distribution of genuine faces and develop a facial match based on the combination of identified facial features to create a latent space representation which will further help in creating a targeted search space for that particular criminal.

Another practical application of our work can be in data augmentation tasks, where current methods rely on approaches pertaining to image space but our work provides a more controllable and vast scheme where we create different samples for training set by searching for user defined standard deviation around the mean latent space vector. This gives a guided control search aspect to data augmentation where we can easily fill the gaps of class imbalance and also regularize our model.

\textbf{Negative impact} A vulnerability exploration is just exploitation of the current system and hence our work can be used to develop fooling mechanisms which may fool the face recognition systems or may also help in upscaling the deep fake \citep{Nguyen2019DeepLF} architectures where combination of multiple images will still output realistic rendition, as it happens in our case.

\textbf{Ethics}  From ethics point of view, our work lies in a morally correct area, because as per morals any vulnerability when reported can be fixed and thereby help in making the systems more secure and robust against attacks. Therefore, we consider our work to be morally correct and catering to the societal benefit.



\bibliographystyle{plain}
\bibliography{neurips_2020}

\end{document}